\documentclass[sigconf]{acmart}

\usepackage{booktabs} 
\usepackage{mathrsfs,amsmath}

\setcopyright{rightsretained}

\acmDOI{10.475/123_4}

\acmISBN{123-4567-24-567/08/06}

\acmConference[SIGKDD'17]{Workshop on Machine Learning for Creativity}{Aug 2017}{Halifax - Canada} 
\acmYear{2017}
\copyrightyear{2017}

\begin{document}
\title{A Machine Learning Approach for Evaluating Creative Artifacts}


\author{Disha Shrivastava$^\dagger$  Saneem Ahmed CG$^\dagger$  Anirban Laha$^\dagger$  Karthik Sankaranarayanan$^\dagger$\\
$^\dagger$IBM Research India,\\
\{dishriva,saneem.cg,anirlaha,kartsank\}@in.ibm.com}

\renewcommand{\shortauthors}{D.  Shrivastava et al.}

\begin{abstract}
Much work has been done in understanding human creativity and defining measures to evaluate creativity. This is necessary mainly for the reason of having an objective and automatic way of quantifying creative artifacts. In this work, we propose a regression-based learning framework which takes into account quantitatively the essential criteria for creativity like novelty, influence, value and unexpectedness. As it is often the case with most creative domains, there is no clear ground truth available for creativity. Our proposed learning framework is applicable to all creative domains; yet we evaluate it on a dataset of movies created from IMDb and Rotten Tomatoes due to availability of audience and critic scores, which  can be used as proxy ground truth labels for creativity. We report promising results and observations from our experiments in the following ways : 1) Correlation of creative criteria with critic scores, 2) Improvement in movie rating prediction with inclusion of various creative criteria, and 3) Identification of creative movies.
\end{abstract}

\keywords{Computational Creativity, Movies, Machine Learning}




\maketitle

\section{Introduction}

Humans have been creative since time unknown. The progress of civilizations has happened as our ancestors have found creative ways to solve problems. Creativity is evident in every aspect of human advancement ranging from science to various forms of art. In fact, creative intelligence is considered by many as the ultimate goal of artificial intelligence. The achievements of artificial intelligence in solving real-world problems has been nothing short of extraordinary. However, the ability to generate creative content is one area where much is left desired from machines. Availability of creative content in various domains like music, painting, dance, fashion, literature, movies, games and cooking has fueled research in an emerging field known as \emph{computational creativity}. 

According to \cite{maher2010}, research in computational creativity is motivated by the following goals: first, modelling the creative generation process to enable machines to generate creative content; second, enabling machines to augment human creativity, and third, assessing and identifying creativity without any assumption of an artifact being produced by machines or humans. Understanding creativity entails the following: first, quantitative evaluation of creativity based on different criteria, and second, learning creative scoring models on top of above criteria. Following this idea, we formulate an automated learning based approach to evaluate creative artifacts. 

According to the recent literature \cite{elgammal15,regent16,maher2010,maher2013,Mariani2016IdentificationOM,varshney13,varshneyicci2013}, the essential criteria for creativity are mainly the following: \emph{value}, \emph{novelty}, \emph{influence} and \emph{unexpectedness}. 
\emph{Value} of an artifact is a measure of how well it is perceived by the society in isolation; without any comparison with other artifacts in its domain. That is, how appealing or useful or pleasant or performing the artifact is; the exact criterion being heavily dependent upon the domain and the users. As an example, one user can rate the movie `Silence of the Lambs' high because of good acting and direction, whereas another user can rate it low because of the presence of violent and disturbing stills in the movie. \emph{Novelty} of an artifact is a notion of how different it is from other artifacts in its domain. For example, `Honey Butter Pepper Lemon BBQ Chicken' is a novel recipe, as nobody has tried it out, even though it may taste weird and hence, not valuable. \emph{Influence} of an artifact quantifies how much impactful or inspiring it has been towards artifacts occurring later in time. For example, `Mona Lisa' is known to have been an inspiration for Raphael's painting `Young Woman with Unicorn' and many others. Another criteria is the \emph{unexpectedness} or \emph{surprise} which defines how different the artifact or some of its attributes are from expected behavior. For example, a black-and-white movie like `The American Astronaut' in year 2001 is unexpected, even though such movies are not novel as they have appeared in the past. 



Convinced with the idea that evaluation of creative artifacts should involve all the above four criteria, we proceed to build a machine learning framework incorporating them; to come up with an objective and automatic way of evaluating creativity which will be applicable to all domains. In this approach, we first compute the criteria-specific measures (value, novelty, influence and unexpectedness), which can be then used as features for a regression-based model. From the above definitions, it is clear that all criteria except \emph{value} require comparison of an artifact with others with the help of some similarity function. The similarity function can vary based on the domain. Our framework assumes the availability of a similarity function to define the computation of measures for different creative criteria to be included in the regression model. Hence, this approach is applicable to all domains, conditional on the availability of similarity function for the corresponding domains.


Even though our approach can be used to evaluate creative artifacts in all domains, we choose to perform our experiments on the movies dataset due to the availability of ground truth ratings like critics score and audience score. According to \cite{lowagreement17}, \emph{audience ratings correlate only weakly with the judgment of professional movie critics}. Although the movie `Moonlight' was hailed by critics, it was not received very well by audience. On the other hand, `Interstellar' was slammed by critics even though it was a huge success. This is perhaps because the critics are well-versed with the landscape of movies, and hence give a higher importance to novelty and influence than the audience. We justify the above claim in our experiments by showing that measures for those criteria are highly correlated with critics score. In our prediction model, the inclusion of various creative criteria lead to improved performance in predictions in both audience and critics score. Previous work by \cite{debarun16} has tried to combine multiple criteria to evaluate artifacts using various heuristics. To the best of our knowledge, this is the first machine-learning approach which uses ideas from philosophies of creativity to come up with a rating prediction model.



In a nutshell, our contributions are primarily the following:
\begin{enumerate}
    \item A domain-independent learning based framework incorporating measures for novelty, influence, value and unexpectedness for evaluating creative artifacts.
    \item Empirical analysis showing that incorporating various creative criteria improves the rating prediction of movies.
    \item The above measures are highly correlated to objective measures like critics score compared to audience score.
\end{enumerate}


\section{Proposed Framework}
In this section, we are going to elaborate the details of the learning based framework that can be used to evaluate creativity of artifacts in any domain. Let us consider a set of $m$ artifacts $\{M_i\}^{m}_{i=1}$ which we want to rate based on creativity. Based on the information $a_i$ available about the artifact $M_i$ , the artifact can be rated as:

\vspace*{-5mm}
\begin{align}
    y_i = \Phi (a_i) \label{eqn1}
\end{align}

Here, $\Phi$ is a function which takes all the information available about the artifact as input and produces the rating $y_i$. Here, $a_i = \{a_{i1},...,a_{iN}\}$ corresponds to a list of $N$ attributes. Examples of attributes can be \emph{genre, runtime, budget, plot}, etc in case of movies. For a fashion apparel, attributes can be \emph{gender, apparel type, brand, colour, texture, image of the apparel}, etc. Function $\Phi$ can be broken down further into a feature extraction step followed by prediction step as follows:
\vspace*{-4mm}
\begin{align}
    \textbf{x}_i = \mathscr{F} (a_i) \\ \textbf{x}_i = [\textbf{f}_i
    ,\textbf{n}_i,\textbf{l}_i,\textbf{u}_i]\\
    y_i = \Psi (\textbf{x}_i)
\end{align}

The function $\mathscr{F}$ involves extraction of different kinds of features like \emph{value} features $\textbf{f}_i$,
\emph{novelty} features $\textbf{n}_i$, \emph{influence} features $\textbf{l}_i$ and \emph{unexpectedness} features $\textbf{u}_i$ from $a_i$. Thus, given a set of $m$ datapoints (or artifacts) in the form $S = \{a_i,y_i\}_{i=1}^{m} $, the function $\mathscr{F}$ is used to transform them to training data $D = \{\textbf{x}_i,y_i\}_{i=1}^{m} $ for the regression model $\Psi$, which takes input feature $\textbf{x}_i$ for artifact $M_i$ to produce creativity rating $y_i$. 



\subsection{Value}
\label{sec:value}
Previous works like \cite{maher2010,regent16,maher2013,varshney13,varshneyicci2013} have highlighted that \emph{value} is an important criteria for creativity. In \cite{maher2010}, value is defined as \emph{a measure of how the artifact is valued by domain experts for this class of artifact or is a reflection of the acceptance of this artifact by society}. For example, a movie directed by `Christopher Nolan' will be perceived as highly valuable compared to a lesser known director. This fact is elucidated very well in \cite{regent16} for the domain of fashion, where certain colour combinations of apparels are valued higher based on a synergy list consisting of most prevalent colours. 

Every artifact $M_i$ can have $N$ attributes $ \{a_{i1},...,a_{iN}\}$, which could be of various types, like numerical, categorical, textual and even different modality like image, etc. Numerical attributes need not have any encoding and their values can be used as is, whereas categorical attributes need one-hot encoding and textual or other modalities require embedding-based representations. Application of encoding schemes and dimensionality reduction technique like Principal Components Analysis (PCA) can be applied to produce a list of vectors $\textbf{f}_i = \{\textbf{f}_{i1},...,\textbf{f}_{iN}\}$. This list of vectors $\textbf{f}_i$ can be treated as an input vector (produced by feature extraction from $a_i$) to a regression model $\Theta$ to predict the output rating $y_i$:

\vspace*{-5mm}
\begin{align}
    y_i = \Theta (\textbf{f}_i) \label{eq2}
\end{align}

In the equation ~\ref{eq2} above, the learning model tries to assign higher weights to dimensions of $\textbf{f}_i$ which are highly correlated with the output rating. So, the model will consider a datapoint highly valuable if its combination of correlated dimensions leads to a better rating prediction. In our formulation, the \emph{value} aspect of creativity is inherently captured by the regression-based model $\Theta$ when the features for every datapoint $i$ is extracted in isolation without considering other datapoints.


\subsection{Novelty and Influence}\label{sec:nov_inf}
In our framework, we propose to include two important metrics computed from the similarity graph obtained from the data points : \emph{novelty} and \emph{influence}. According to \cite{maher2010}, novelty is \emph{a measure of how different the artifact is from known artifacts in its class}. In other words, if an artifact is far apart from other known artifacts, then it is considered highly novel. For example, the first ever time time-travel movie will have high novelty score compared to other time-travel movies which came much later. Similarly, there can be artifacts which can be completely different from other known artifacts, hence novel, yet missing out on the value aspect.

On the other hand, \emph{influence} measures if an artifact was an inspiration or if it had an impact on another artifact. For example, `The Godfather' movie had significant influence on its sequel. Similarly, the first time-travel movie would have had significant influence on similar movies (`The Terminator') appearing later in time. In the domain of fashion, designers tend to get inspired by works of other designers, which means the previous works had an influence on the later ones. It is also, however, possible that non-novel artifacts can also influence later works.

As is evident from the above definitions, novelty and influence measures for an artifact cannot be computed in isolation from other artifacts as they require similarity with other artifacts. It is to be noted that even though a design appearing later in time may not have been directly inspired by an earlier one, but in the eyes of the observer, the earlier one is influential if there is a high similarity. Considering the observer/critic point of view (in line with Boden's view of H-creativity \cite{boden91}), these measures are highly related to the chronological order of the artifacts, as the earlier artifacts tend to have influence over the later ones. Some works like \cite{gravino16,Wasserman03022015,andreas14} have already mentioned using chronological order of movies and measuring influence based on a movie citation network which includes explicit references between movies. Explicit references lead to very sparse connections and may have incomplete information as they are manually curated by IMDb. Hence, we suggest using a similarity based graph for artifacts as used in  \cite{Mariani2016IdentificationOM} for identifying milestone papers, in \cite{elgammal15} for quantifying creative paintings and in PageRank\cite{Pageetal98} for calculating importance of web documents.

The first step towards construction of a similarity graph based analysis is computation of the similarity matrices $\textbf{W}_1,...,\textbf{W}_N$ between all pairs of artifacts corresponding to each attribute $k$ of $N$ attributes. The type of attribute dictates the choice of similarity measure as follows: 
\begin{align}
    Sim(M_i,M_j,k) =
    \begin{cases}
        lin(f_{ik}, f_{jk}) = \frac{1}{1+|f_{ik} - f_{jk}|}, \text{numerical}\ k.\\
        \\
        exp(f_{ik}, f_{jk}) = e^{-|f_{ik} - f_{jk}|}, \text{numerical}\ k.\\
        \\
        cos(\textbf{f}_{ik}, \textbf{f}_{jk}) = \frac {\textbf{f}_{ik} \cdot \textbf{f}_{jk}}{||\textbf{f}_{ik}|| \cdot ||\textbf{f}_{jk}||}, \text{others.}
    \end{cases} \label{eq:sim}
\end{align}

Every artifact $M_i$ also comes with a time of occurrence $t(i)$ based on which a selection matrix \textbf{S} is computed. The main idea here is to have a graph among the artifacts, the edges being related to the similarity between the nodes, in such a way that the novelty and influence scores of the artifacts can be computed using PageRank\cite{Pageetal98} algorithm. This follows from the logic that if there is a high similarity edge between two nodes, the source node is more influential and novel than the target and hence, more creative. The complete sequence of steps (refer \cite{elgammal15} for details) to be followed for computing the parameters of the graph are listed below:
\begin{align}
    w_k(i,j) = Sim(M_i,M_j,k),\ \forall\ i,j,k \label{sim}\\
    s(i,j) = 
    \begin{cases}
        1, & \text{if}\ t(i)<t(j)  \\
        0, & \text{otherwise}
    \end{cases} \label{sel1}
    \\
    w_k(i,j) = w_k(i,j) * s(i,j),\ \forall\ i,j,k \label{sel2}\\
    \hat{w}_k(i,j) = w_k(i,j) - \tau_k,\ \forall\ i,j,k \label{thresholding}\\
    Reversal : 
    \begin{cases}
        \hat{w}_k(i,j) \ge 0 \rightarrow \tilde{w}_k(i,j) = \hat{w}_k(i,j) \\
        \hat{w}_k(i,j) < 0 \rightarrow \tilde{w}_k(j,i) = -\hat{w}_k(i,j)
    \end{cases},\
    \forall\ i,j,k \label{rev}\\
    p(i,j) =
    \begin{cases}
        1, & \text{if}\ t(i)>t(j)  \\
        0, & \text{otherwise}
    \end{cases} \label{sel3}
    \\
    \tilde{w}^p_k(i,j) = \tilde{w}_k(i,j) * p(i,j)\ \forall\ i,j,k \label{sel4}
    \\
    q(i,j) =
    \begin{cases}
        1, & \text{if}\ t(i) \le t(j)  \\
        0, & \text{otherwise}
    \end{cases} \label{sel5}
    \\
    \tilde{w}^q_k(i,j) = \tilde{w}_k(i,j) * q(i,j)\ \forall\ i,j,k \label{sel6}
\end{align}

Eqn ~\ref{sim} is related to the similarity matrices $\textbf{W}_k$ created for every attribute $k$. Similarity matrices being symmetric, the graph formed by $\textbf{W}_k$ is an undirected graph. The next steps in equations ~\ref{sel1} and ~\ref{sel2} select only those edges in the graph which go forward in time, which is also the direction of influence, thus producing a directed graph. This is followed by a thresholding step in equation ~\ref{thresholding}, which mandates only high value edges eligible for score propagation.
This step leads to addition of more negative weighted edges to the graph, which can be corrected by an edge reversal step in equation ~\ref{rev}, resulting in creation of edges not going forward in time. However, to preserve the distinction between edges which were reversed and the others, the graph is split into two directed graphs, \emph{prior} graph and \emph{subsequent} graph in the remaining steps. As elaborated in \cite{elgammal15}, the former graph leads to computation of novelty scores whereas the latter leads to generation of influence scores. Equations ~\ref{sel4} and ~\ref{sel6} along with column normalization step produces two column-stochastic matrices $\tilde{\textbf{W}}^p_k$ and $\tilde{\textbf{W}}^q_k$ for every attibute $k$. These matrices together help in computing the \emph{novelty} scores $n_{ik}$, \emph{influence} scores $l_{ik}$ and \emph{aggregate} scores $c_{ik}$ based on the following equations:
\begin{align}
    \textbf{c}_k = \frac{1-\alpha}{m}\textbf{1} + \alpha [\beta \tilde{\textbf{W}}^p_k \textbf{c}_k + (1-\beta) \tilde{\textbf{W}}^q_k \textbf{c}_k] \label{calc}\\
    \Rightarrow \textbf{c}_k = [\frac{1-\alpha}{m}\textbf{E} + \alpha \beta \tilde{\textbf{W}}^p_k + \alpha (1-\beta) \tilde{\textbf{W}}^q_k] \textbf{c}_k,\\
    \text{where}\ \textbf{E}\ \text{is all 1's matrix.} \nonumber \\
    \Rightarrow \textbf{c}_k = \mathcal{M} \textbf{c}_k, \text{where} \mathcal{M}\ \text{is column stochastic.} \label{eigen}\\
    \Rightarrow \textbf{n}_k = \alpha \beta \tilde{\textbf{W}}^p_k \textbf{c}_k\ \text{,}\ \textbf{l}_k = \alpha (1-\beta) \tilde{\textbf{W}}^q_k \textbf{c}_k \label{novinf}
\end{align}
The power iteration method \cite{Pageetal98} can be used to solve the above equation ~\ref{eigen}. Now, substituting back the solution $\textbf{c}_k$ in equation ~\ref{calc}, we get the novelty scores $\textbf{n}_k$ and $\textbf{l}_k$ over all $m$ artifacts for attribute $k$ as in equation ~\ref{novinf}. On computation of above for all attributes results in instance specific scores $\textbf{n}_i$, $\textbf{l}_i$, and $\textbf{c}_i$ for artifact $M_i$, each of these vectors being of same dimension as number of attributes.


\subsection{Unexpectedness}\label{sec:unex}
\emph{Unexpectedness}, also known as \emph{surprise} is defined as a criteria for creativity, which measures if an artifact is different from its expected attributes, the amount of difference may not be important \cite{maher2010}. Typically,  the evaluation of unexpectedness considers only the recent past to figure out if an artifact or its attributes is unexpected. For example, a black-and-white movie in 1920s will not be unexpected as colour movies were not yet the norm. However, the 2011 French movie `The Artist' would have been considered highly unexpected in the age of colour movies. Another curious example is that from wide-rimmed glasses, which were pretty much in fashion till 1990s and for a brief period went out of fashion. So, if such an artifact appeared in early 2000s, it would have been highly unexpected compared to current years as they are back in fashion again. There is a fine difference between novelty and unexpectedness as novelty implies unexpectedness but not the other way round. The unexpectedness defined here is different from Boden's notion of P-creativity \cite{boden91} as our notion considers if an artifact is unexpected at a certain time rather than if it is unexpected from its creator.

In this work, we measure unexpectedness by considering the recent past (say, 5 years) of an artifact $M_i$ and how different it is from its predecessors in the recent past, belonging to the set $\gamma = \{j: t(i-5) \le t(j) < t(i)\}$. Here, for each attribute $k$, the unexpectedness of the artifact is defined separately as $u_{ik}$ by the following different measures as below:

\begin{align}
    u_{ik} = 
    \begin{cases}
        max\_unexp(i,k) = - \underset{j}{\max}\ Sim(M_i,M_j,k)\\
        avg\_unexp(i,k) = - \frac{1}{|\gamma|} \underset{j}{\sum}\ Sim(M_i,M_j,k)\\
        inv\_unexp(i,k) = - \frac{\underset{j}{\sum} v_{ij} * Sim(M_i,M_j,k)}{\underset{j}{\sum} v_{ij}},\\ \ \ \ \text{where}\ v_{ij} = \frac{1}{|t(i)-t(j)|}
    \end{cases} \label{unexp}
\end{align}

All of the measures rely on the fact that the more similar an artifact $M_i$ is to its recent predecessors, the less unexpected it will be. The different measures considered are maximum similarity, mean similarity and inverse weighted mean similarity, where the weights are higher for a recent movie compared to an earlier one. Thus, for every artifact $M_i$, unexpectedness scores $\textbf{u}_i$ can be generated by one of the above measures in equation ~\ref{unexp}, where the dimension of the score vector is the number of attributes. 





\section{Experiments}
For our experiments we have considered a single dataset in the domain of movies but the techniques developed in this paper can be extended to any other creative domain. Experiments were done on 5000+ movies metadata from IMDb and Rottentomatoes (RT) website. Two sets of experiments were carried out. First one analyzes how various \textit{creativity} criteria-specific measures (\textit{novelty}, \textit{influence}, \textit{aggregate} and \textit{unexpectedness}) are correlated with popularity and ratings of these artifacts. Second one is to use these measures in a prediction model to get better accuracy in predicting popularity and rating. 

Rest of this section is divided into two sub-sections. First sub-section explains data preparation and computation of \textit{creativity} criteria-specific measures. Second section explains details of the analysis carried out on the data including the model used to predict popularity of movie using these measures.

\subsection{Data Preparation and Computing \textit{Creativity} Measures}
\subsubsection{\textbf{Dataset Collection and Curation}}
The Internet Movie Database (IMDb) \footnote{www.imdb.com} and RottenTomatoes (RT) \footnote{www.rottentomatoes.com} are public websites with a large collection of metadata for movies, television programs and other visual media. IMDb users are invited to rate any film on a scale of 1 to 10, and the totals are converted into a weighted mean-rating that is displayed beside each movie title. Similarly in Rotten Tomatoes, each movie has critic ratings/scores and audience ratings/scores. The reviews from top movie critics are collated to form a critic rating/score. 
We obtained 28 attributes for 5043 movies from Kaggle IMDb 5000 Movie Dataset \cite{chuansun2016}. In addition, we scraped additional attributes for these movies from the RT website and corresponding Wikipedia movie plots to get a total of 49 attributes. Out of these, some attributes like  `director\_name', `actor\_1\_name', `actor\_2\_name', `actor\_3\_name', `movie\_imdb\_link', 'movie\_release\_date', `movie\_writer\_name', etc. were dropped either because they conveyed no useful information or the information they conveyed can be expressed by some other attribute e.g. director\_facebook\_likes was more useful than the name of the director for our analysis. We also dropped attributes like box-office collection of a movie, number of audience/critics who rated the movie, number of facebook likes of the movie, etc. which are heavily correlated to the rating of the movie. Following attributes were used as output labels to be predicted by ML models:
\begin{enumerate}
    \item \textsc{IMDb rating} (1-10)
    \item \textsc{RT Audience rating} (1-5)
    \item \textsc{RT Audience score} (0-100)
    \item \textsc{RT Critics rating} (1-10)
    \item \textsc{RT Critics score} (0-100)
\end{enumerate}

Computation of measures for some criteria requires time when an artifact was created. We removed movies which had their release year missing. If either of IMDb year of release or RT year of release for the movie were present, we kept that movie entry. If both the entries were present, we take the IMDb year. If none were present, we discard that movie. After this elimination, we were left with 5021 movies to proceed further.

\subsubsection{\textbf{Attribute Representation}}
After the curation of the dataset, we finally had 21 unique attributes (Table \ref{t1}). In the Table, we also report the number of missing values for each attribute. The reason for missing values is either because they were not present in the original Kaggle-dataset or due to inconsistencies in matching movies between RT and IMDb during the web-scraping of movies from the RT website. We filled in the mean of the corresponding attributes and `NA' for missing numeric and string attribute values, respectively. 

Next, we needed to convert categorical attributes to numerical values to be used further in our analysis. For attributes which consisted of single string and had semantic meaning associated with them ( e.g. plot\_keywords, genres, country, language), we used Word2Vec \cite{word2vec} to get vectors of dimension 300 each. The attributes which consisted of one or more sentences (like Wiki\_plot, RT\_plots) were converted to Skip-Thought vectors\cite{kiros15} of dimension 4800 each, by average pooling across sentences. For attribute values which did not have a word2vec or skip-thought representation,we took a zero vector of corresponding size. The vectors for attributes with multiple strings( like genres and plot\_keywords) or multiple sentences (like wiki\_plot) were obtained by taking the mean vector across all strings/ sentences which represent that attribute. Attributes such as color and rt\_studio were converted to one-hot encoded vectors. 

To reduce the large dimensionality of the feature space(\textasciitilde20000) as compared to the number of movie samples(\textasciitilde5000), we applied PCA on individual skip-thought and word2vec vectors such that 90\% of the variance is captured by them. This resulted in 407 features(in total for 21 attributes) which were then used as input for calculation of similarity measures. These features will be called as \textbf{PCA features} in future references. These features have been computed in isolation for an artifact and depict the \emph{value} of the artifact as explained in Section \ref{sec:value}.

\begin{table} 
    \centering
    \resizebox{0.45\textwidth}{!}
    {
    \begin{tabular}{|l|c|c|r|}
        \hline
    \textbf{Attribute} & \textbf{Representation} & \textbf{Similarity} & \textbf{Missing Values} \\ \hline
	IMDb plot & Skip-Thought & Cosine & 2484 \\ \hline
	Wiki plot & Skip-Thought & Cosine & 1339 \\ \hline
	Language & Word2Vec & Cosine & 12 \\ \hline
	Color & One-Hot & Cosine & 19\\ \hline
	Certification & One-Hot & Cosine & 303\\ \hline
	Why certification & Skip-Thought & Cosine & 1442\\ \hline
	RT plot & Skip-Thought  & Cosine & 513\\ \hline
	Studio & One-Hot Encoded  & Cosine& 566 \\ \hline
	Genres & Word2Vec  & Cosine & 0 \\ \hline
	Movie title & Word2Vec  & Cosine & 0 \\ \hline
	Plot keywords & Word2Vec  & Cosine & 153 \\ \hline
	Country & Word2Vec  & Cosine & 5 \\ \hline
	Actor-1 FB likes & Real Number & Linear, Exp & 7 \\ \hline
	Actor-2 FB likes & Real Number & Linear, Exp & 13 \\ \hline
	Actor-3 FB likes & Real Number & Linear, Exp & 23 \\ \hline
	Aspect ratio & Real Number & Linear, Exp & 329 \\ \hline
	Budget & Real Number & Linear, Exp & 492 \\ \hline
	Cast total  FB likes & Real Number & Linear, Exp & 0 \\ \hline
	Director FB likes & Real Number & Linear, Exp & 104 \\ \hline
	Runtime & Real Number & Linear, Exp & 15 \\ \hline
    \# of Faces in poster & Real Number & Linear, Exp & 13\\ \hline  

 \hline
    \end{tabular}}
    \caption{Different Attributes used along with their representation, similarity type and number of missing values}
    \label{t1}
\end{table}

\subsubsection{\textbf{Computing similarity}}
We employed mainly three types of similarity measures depending on the representation of the attribute. For real-valued attributes, we used linear and exponential similarity measures (See Eqn \ref{eq:sim}) and for the rest, we used cosine similarity (Table \ref{t1}). For each attribute, we obtained a $N \times N$ similarity matrix (where $N$ = number of movies) by finding the similarity between all pairs of movie combinations. Before passing the numeric attributes for calculating linear and exponential similarities, we normalized them to avoid scaling issues.

\subsubsection{\textbf{Computing \textit{Creativity} Measures}}

To compute the novelty, influence and aggregate scores we implemented the graph network as discussed in Section \ref{sec:nov_inf} above. \textit{Aggregate} is combination of novelty and influence which is obtained from eqn(\ref{calc}). The movies formed the nodes of the graph. The similarity value between two movies was used as a weight of the edge connecting the two movies in the graph. We obtained one graph for each attribute and similarity type. We used $\alpha = 0.95$, $\beta = 0.2$ and $0.5$ in eqn(\ref{calc}) to compute novelty, influence and aggregate. Experimentally we found that using $\beta=0.2$ gives better results and hence we are only reporting results with $\beta=0.2$.

For calculating mean, max and inverse\_weighted measures of unexpectedness (See Section \ref{sec:unex} for details), we took movies in a 5 year window past the current movie year. Amongst the calculated unexpectedness measures, \emph{mean} was used for feeding into the prediction models.

\subsection{Analysis}
\subsubsection{\textbf{Correlation Analysis}}
We obtained the Pearson Correlation of our attribute-wise influence, novelty, aggregate (of novelty and influence) and unexpectedness scores with each of the output labels. Though we computed aggregate scores, novelty scores and influence scores with $\beta=0.2$ and $\beta=0.5$; we are only reporting results for $\beta = 0.2$. We obtained correlation values for mean, max and inverse\_weighted unexpectedness scores as well. We are only reporting results of max-unexpectedness in this paper for the interest of space.

\subsubsection{\textbf{Prediction Models}}

We experimented with different combinations of attributes and different models for predicting the values of the five output labels. Our aim was to show that using \textit{creativity} measures (novelty, influence, aggregate and unexpectedness) with PCA features gives better results than just using PCA features.

To make experiments across output labels comparable, we normalized them by dividing with their highest possible value (e.g. IMDb ratings were divided by 10) before giving to ML models for training. Experiments were carried out on the following combinations of features:
\begin{enumerate}
    \item \textbf{Baseline}: We consider PCA features as the baseline for our experiments.
    \item \textbf{PN}: PCA features + Novelty measure
    \item \textbf{PI}: PCA features + Influence measure
    \item \textbf{PU}: PCA features + Unexpectedness measure
    \item \textbf{PUN}: PCA features + Unexpectedness measure + Novelty measure
    \item \textbf{PUI}: PCA features + Unexpectedness measure + Influence measure
    \item \textbf{PUNI}: PCA features + Unexpectedness measure + Novelty measure + Influence measure
    \item \textbf{PUNIA}: PCA features + Unexpectedness measure + Novelty measure + Influence measure + Aggregate measure.
\end{enumerate}

We trained five different types of models: Support Vector Regressor, Random Forest Regressor, Ridge Regression, Bayesian Regression and K-Nearest Neighbour Regressor for each combinations above. We used the implementations of these models as present in scikit-learn version 0.18.1\cite{scikit-learn}. For KNN, we used n\_neighbours = 5; and for Random Forest Regressor we used n\_estimators=600 and min\_samples\_split=2. For other models, we used the default values of parameters. We converted the data to zero mean and unit standard deviation before giving to the models for training. We used a train-test split of 80-20 \%. After the training was done, root mean square error (RMSE) was used as a measure of test error.

\begin{figure*}[h]
\raggedleft
\includegraphics[width=2.5\columnwidth]{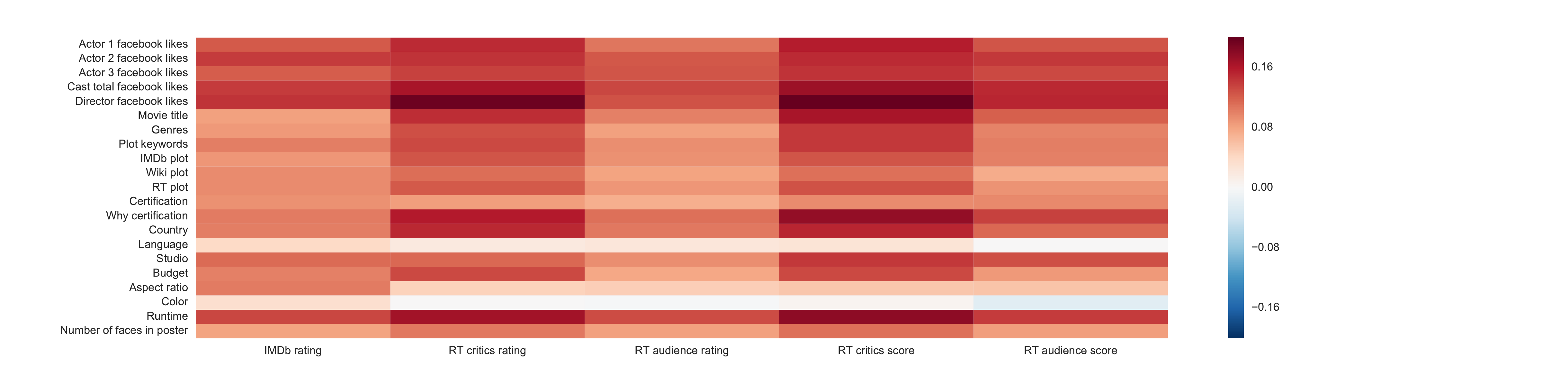}
\caption{Pearson correlation of aggregate of novelty and influence scores of different attributes with labels. }\label{fig:creat1}
\end{figure*}

\begin{figure*}[h]
\raggedleft
\includegraphics[width=2.5\columnwidth]{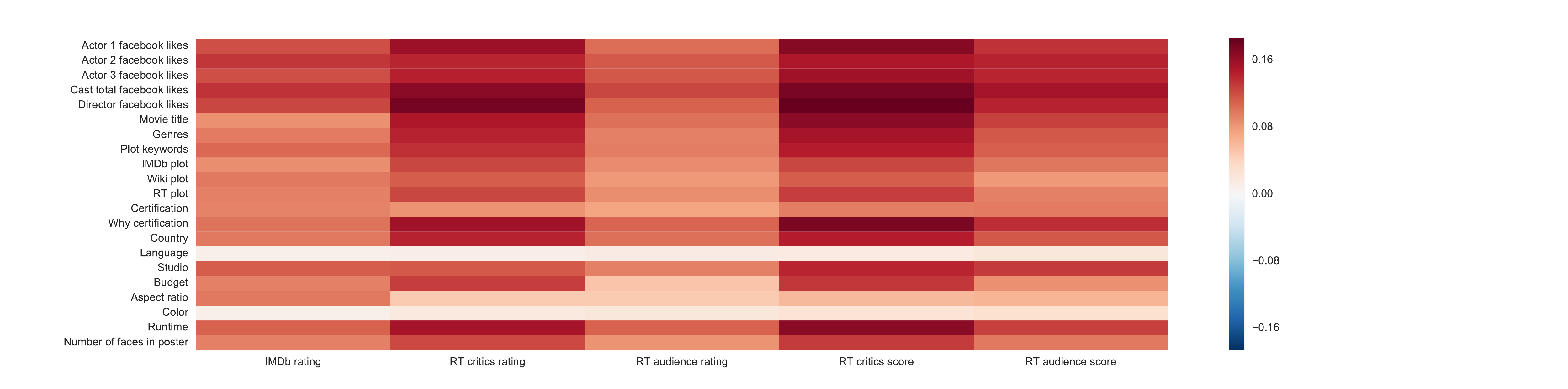}
\caption{Pearson correlation of influence scores of different attributes with labels. }\label{fig:inf1}
\end{figure*}

\begin{figure*}[h]
\raggedleft
\includegraphics[width=2.5\columnwidth]{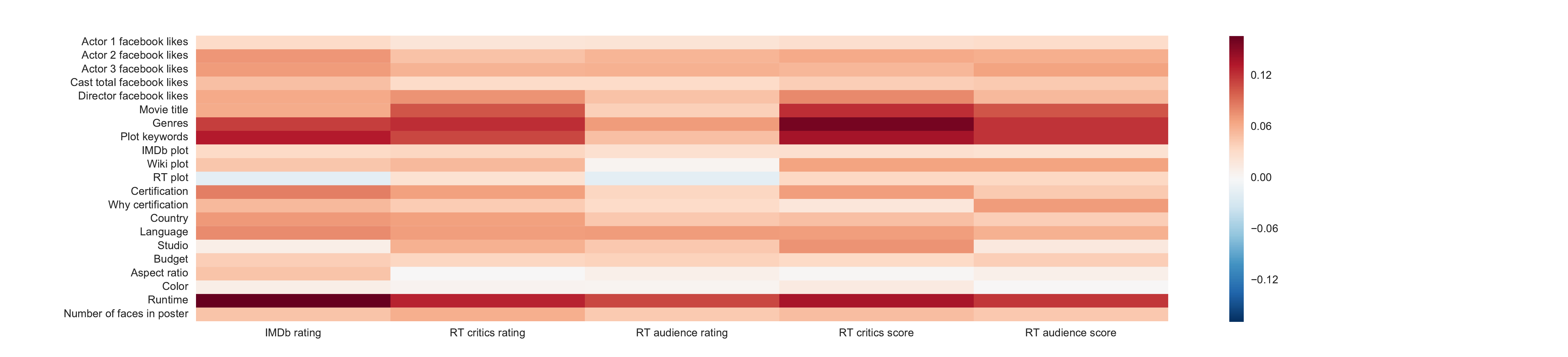}
\caption{Pearson correlation of unexpectedness scores(max) of different attributes with labels. }\label{fig:unex1}
\end{figure*}

\begin{figure*}[h]
\raggedleft
\includegraphics[width=2.0\columnwidth]{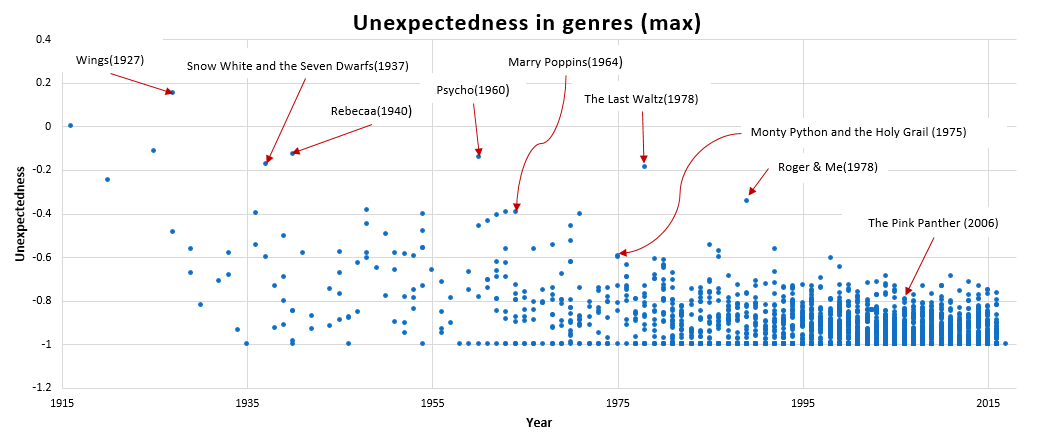}
\caption{Visualization of unexpectedness scores of movies.}\label{fig:unex_scatter}
\end{figure*}

\section{Results and Discussion}
Though experiments were done on the metadata of the movies and not on the actual content of the movie (which is the video and audio of the movie), we were able to find promising observations and results. We will first discuss results from correlation analysis and later explain rating prediction findings.

\subsection{Correlation Analysis}
In the first set of experiments we computed the pearson correlation between 
\textit{novelty}, \textit{influence}, \textit{aggregate} and \textit{unexpectedness} with popularity and ratings of the movies. These measures were computed separately for each attribute. Again \textit{aggregate} here is a combination of novelty and influence which is obtained from eqn(\ref{calc}). 

Figures \ref{fig:creat1}, \ref{fig:inf1} and \ref{fig:unex1} shows the correlation heatmap obtained for \textit{aggregate}, \textit{influence} and \textit{unexpectedness} measures, respectively. We were able to make following observations.
\begin{enumerate}
    \item From the heatmaps we can see that \textit{creativity measures} are almost all positively correlated with the different ratings. This confirms the hypothesis that these \textit{measures} can indeed capture audience and critics interests.
    \item In figure \ref{fig:creat1} and \ref{fig:inf1} we can see that second and fourth column are more darker than other columns which implies that \textit{aggregate} and \textit{influence} measures are more correlated with critic rating and critic score compared to audience rating and score. This might point out that critics give more importance to influence and novelty of a movie compared to audience.
    \item \textit{Plot keywords} are more correlated to ratings compared to detailed plots of the movie: \textit{IMDb plot, wiki plot, RT plot}. Inability of skip-thought vectors to represent a large text content might be the reason for this. Among the plots, \textit{wiki plot} is the longest and detailed plot in our dataset. We could see that correlation values for \textit{wiki plot} is the least among plots. More texts might have diluted the main content of the plot in the representation.
    \item \textit{Language}, \textit{Color} and \textit{Aspect ratio} are least correlated attribute in all three \textit{measures}. 
\end{enumerate}

From figure (\ref{fig:unex1}) one could see that unexpectedness on \textit{Genres} is highly correlated with the RT critic score. To further understand how unexpectedness scores were given to each movie in this case, we obtained a scatter plot in figure(\ref{fig:unex_scatter}). It is very interesting to see many famous movies getting higher scores in the figure. One could see that "Snow White and The Seven Dwarfs(1937)", first color animation movie with voice has got a high score. Similarly "Psycho(1960)" ranked among most greatest films has also got high score. These analyses further confirm that using \textit{unexpectedness} and other \textit{creativity} criteria on creative artifacts is the right direction to pursue.

\subsection{Prediction Task}
Here we report results of experiments ran on three output labels: \textsc{IMDb rating}, \textsc{RT Critic Score} and \textsc{RT Audience Score} (skipping others for the interest of space as they have similar observations). In Tables \ref{t2}-\ref{t4}, we present the RMSE on the test data for different combinations of datasets and models with each output label. The best score for each model has been highlighted in red and the best percentage increase across all models and datasets has been highlighted with blue in the Tables. As can be clearly seen from the numbers, the inclusion of \textit{influence}, \textit{novelty}, \textit{aggregate} and \textit{unexpectedness} scores leads to better prediction performance for all models and output labels.
In all cases, better performance from the baseline is shown. This shows that creativity measures are important features for our prediction model. Tables show that we are getting improvement as high as 4.5\% in RMSE.

\begin{table*} 
    \centering
    \resizebox{0.8\textwidth}{!}
    {
    \begin{tabular}{|l|r|r|r|r|r|}
        \hline
    \textbf{Combination} & \textbf{Ridge Regression} & \textbf{Bayesian Regression} & \textbf{SVR} & \textbf{KNN} & \textbf{Random Forest} \\ \hline
Baseline & 0.08983 & 0.08935 & 0.09402 & 0.11224 & 0.09042 \\ \hline
PN & 0.0891 & 0.0886 & 0.09274 & 0.11004 & 0.08929 \\ \hline
PI & 0.09182 & 0.08942 & 0.09365 & 0.11148 & 0.08989 \\ \hline
PU & 0.08931 & 0.08895 & 0.09308 & 0.10963 & 0.08953 \\ \hline
PUN &0.08896 & \textcolor{red}{\textbf{0.0882}} & \textcolor{red}{\textbf{0.09224}} & 0.10777 & 0.08908 \\ \hline
PUI &0.08968 & 0.0889 & 0.09289 & 0.10877 & 0.08922 \\ \hline
PUNI &\textcolor{red}{\textbf{0.08826}} & 0.08871 & 0.0931 & 0.10835 & 0.08936\\ \hline
PUNIA & 0.0898 & 0.08844 & 0.09232 & \textcolor{red}{\textbf{0.1071}} & \textcolor{red}{\textbf{0.08891}} \\ \hline
\textbf{Improvement\%} & \textbf{1.75309
} & \textbf{1.29546
} & \textbf{1.88974
} & \textcolor{blue}{\textbf{4.58040}
} & \textbf{1.67013
} \\ \hline
	
    \end{tabular}}
    \caption{Comparison of RMSE on different combinations and different models for output label - \textsc{IMDb rating}}
    \label{t2}
\end{table*}

\begin{table*} 
    \centering
    \resizebox{0.8\textwidth}{!}
    {
    \begin{tabular}{|l|r|r|r|r|r|}
        \hline
    \textbf{Combination} & \textbf{Ridge Regression} & \textbf{Bayesian Regression} & \textbf{SVR} & \textbf{KNN} & \textbf{Random Forest} \\ \hline
Baseline & 0.24125 & 0.23533 & 0.23151 & 0.26919 & 0.23472 \\ \hline
PN & 0.24095 & 0.23518 & 0.23107 & 0.26899 & 0.23119 \\ \hline
PI & 0.24673 & 0.23615 & 0.22868 & 0.26955 & 0.23097 \\ \hline
PU & 0.24104 & 0.23546 & 0.23065 & 0.26639 & 0.23101 \\ \hline
PUN &\textcolor{red}{\textbf{0.24038}} & \textcolor{red}{\textbf{0.23453}} & 0.22989 & 0.26782 & \textcolor{red}{\textbf{0.23040}} \\ \hline
PUI &0.24742 & 0.23652 & 0.22838 & 0.26776 & 0.23042 \\ \hline
PUNI & 0.24350 & 0.23483 & 0.22806 & 0.26952 & 0.23136 \\ \hline
PUNIA & 0.24799 & 0.23609 & \textcolor{red}{\textbf{0.22765}} & \textcolor{red}{\textbf{0.26539}} & 0.23127 \\ \hline

\textbf{Improvement\%} & \textbf{0.35934
} & \textbf{0.34056
} & \textbf{1.66685
} & \textbf{1.41486
} & \textcolor{blue}{\textbf{1.8394}
} \\ \hline
	
    \end{tabular}}
    \caption{Comparison of RMSE on different combinations and different models for output label - \textsc{RT Critic Score}}
    \label{t3}
\end{table*}

\begin{table*} 
    \centering
    \resizebox{0.8\textwidth}{!}
    {
    \begin{tabular}{|l|r|r|r|r|r|}
        \hline
    \textbf{Combination} & \textbf{Ridge Regression} & \textbf{Bayesian Regression} & \textbf{SVM} & \textbf{KNN} & \textbf{Random Forest} \\ \hline
Baseline & 0.1695 & 0.165 & 0.16259 & 0.20086 & 0.16583 \\ \hline
PN & 0.1684 & 0.16427 & 0.16114 & 0.19898 & 0.16439 \\ \hline
PI & 0.17877 & 0.165 & 0.1614 & 0.19889 & 0.16384 \\ \hline
PU & 0.16881 & 0.16411 & 0.16056 & 0.19659 & 0.16391 \\ \hline
PUN &\textcolor{red}{\textbf{0.16742}} & \textcolor{red}{\textbf{0.16306}} & 0.15954 & 0.19541 & 0.16410 \\ \hline
PUI &0.17528 & 0.16477 & 0.15973 & 0.19451 &\textcolor{red}{\textbf{ 0.16307}} \\ \hline
PUNI & 0.17251 & 0.16529 & 0.15947 & 0.19198 & 0.16342 \\ \hline
PUNIA & 0.17295 & 0.16422 & \textcolor{red}{\textbf{0.1591}} & 0.19332 & 0.16396 \\ \hline

\textbf{Improvement\%} & \textbf{1.22812
} & \textbf{1.17284
} & \textbf{2.1457
} & \textcolor{blue}{\textbf{4.4213}
} & \textbf{1.66459
} \\ \hline

    \end{tabular}}
    \caption{Comparison of RMSE on different combinations and different models for output label - \textsc{RT Audience Score}}
    \label{t4}
\end{table*}

Though this improvement were not as high as we expected, these results are encouraging. It is observed that the percentage improvement in performance over the baseline for audience-related output labels (\textsc{IMDb rating}, \textsc{RT Audience Score}) is better as compared to that for critics-related output labels(\textsc{RT Critic Score}). More investigation is required into this observation. As per the best performing feature combination and ML model are concerned, they vary depending on the output label. Though in most cases, the feature combination with PCA features, unexpectedness and one of the other creativity scores stands out to be the winner. 

A thorough analysis is needed to understand how to extract the best out of these models. Choosing better similarity measures would also help in getting better accuracy. For predicting movie ratings and popularity we feel that there is a limit to which metadata can capture. Using main content of movies (video and audio) should give better accuracy as the ratings/scores are given by the audience and the critics by watching the main content. Among other domains, specifically fashion seems to be the right fit for these techniques as whole or most of the information about the artifact will be captured in an image.

\section{Conclusion and Future Work}
This paper presented a novel machine learning approach to evaluate creative artifacts. We used ideas from philosophies on creativity to build a prediction model which predicts user and critics ratings better. The improvements obtained from our experiments even with movie metadata instead of actual content confirms that this paper is in the right direction. From the correlation analysis, we found that the measures for different creative criteria are more aligned to the critic ratings as compared to the audience ratings. This is expected as critics will consider creativity more while rating a movie as compared to audiences who are also affected by other biases like favouritism towards a particular actor, director, genre and demography. We also observed that our unexpectedness scores were able to come up with reasonable movies at the top. We feel that having good similarity measures can make the \textit{creativity} scores more reliable and might result in better prediction model. We propose to use \textit{metric learning} to find better similarity measure as future extension to this paper. Inclusion of more domains like fashion and music also seem natural extensions as our framework is domain-independent.

\bibliographystyle{ACM-Reference-Format}
\bibliography{sigproc} 

\end{document}